\documentclass[letterpaper]{article} 
\usepackage{aaai2026}  
\usepackage{times}  
\usepackage{helvet}  
\usepackage{courier}  
\usepackage[hyphens]{url}  
\usepackage{graphicx} 
\usepackage{natbib}  
\usepackage{caption} 
\usepackage{algorithm}
\usepackage{algorithmic}
\usepackage{siunitx}
\usepackage{newfloat}
\usepackage{listings}
\DeclareCaptionStyle{ruled}{labelfont=normalfont,labelsep=colon,strut=off} 
\usepackage{amssymb}   
\usepackage{times}     
\usepackage{makecell}   
\usepackage{multirow}   
\usepackage{array}      
\usepackage{microtype}
\usepackage{url}
\usepackage{booktabs}
\usepackage{graphicx}
\usepackage{subcaption}
\usepackage{lineno}
\usepackage{amsmath} 
\usepackage{fancyvrb}
\usepackage{listings}

\floatstyle{ruled}
\newfloat{listing}{tb}{lst}{}
\floatname{listing}{Listing}
\title{DONOD: Efficient and Generalizable Instruction Fine-Tuning for LLMs via Model-Intrinsic Dataset Pruning}

\author {
    Jucheng Hu\textsuperscript{\rm 1,\rm 2}\equalcontrib,
    Suorong Yang\textsuperscript{\rm 1}\equalcontrib,
    Lijun Wu\textsuperscript{\rm 1},
    Dongzhan Zhou\thanks{Corresponding author}\textsuperscript{\rm 1},
}
\affiliations {
    \textsuperscript{\rm 1}Shanghai Artificial Intelligence Laboratory, \\
    \textsuperscript{\rm 2}University College London, \\
    jucheng.hu.20@ucl.ac.uk, sryang@smail.nju.edu.cn, \texttt{\{wulijun,zhoudongzhan\}@pjlab.org.cn}}

\begin{document}

\maketitle

\begin{abstract}
Ad-hoc instruction fine-tuning of large language models (LLMs) is widely adopted for domain-specific adaptation. While domain-specific supervised fine-tuning (SFT) is effective and efficient, it often weakens cross-domain generalization and struggles with noisy training data. To address these challenges, we propose DONOD, a lightweight model-intrinsic data pruning method. Our approach evaluates data using two model-parameter-based metrics: Delta of Norm (DON), which captures the cumulative influence on model weights, and Norm of Delta (NOD), which quantifies weight instability. Moreover, by employing the Technique for Order of Preference by Similarity to Ideal Solution (TOPSIS) algorithm, we effectively filter noisy, unlearnable, and generalization-harming samples without relying on auxiliary models during the SFT process. Experiments on mathematical tasks demonstrate that data selected by DONOD achieves superior fine-tuning efficiency and improved robustness against noisy data. By filtering out 70\% of the whole dataset, we improve target-domain accuracy by 14.90\% and cross-domain accuracy by 5.67\%. Meanwhile, our selected data present superior cross-architecture generalization. Data pruned by smaller models (e.g., Llama 3.1-8B) generalize effectively on larger models (e.g., Llama 2-13B). Compared to existing related methodologies, DONOD demonstrates comparable or superior performance while remaining dataset-agnostic, enabling broader applicability. Code will be made publicly available.
\end{abstract}


\section{Introduction}
In recent years, large language models (LLMs) have demonstrated strong generalization capabilities and remarkable success across a wide range of applications~\cite{achiam2023gpt,meta2024llama32blog,li2024quantityqualityboostingllm}. To further align LLMs with human intent and diverse user needs, supervised fine-tuning (SFT) on curated instruction–response pairs has become a widely adopted approach~\cite{cao2023instruction,2023arXiv230403277P}. However, fine-tuning LLMs with massive instruction datasets incurs substantial computational costs. More importantly, recent studies have pointed out that the quality of the fine-tuning data outweighs its quantity~\cite{li2025scardataselectionstyle, xia2024lessselectinginfluentialdata,wang2023far}. In practice, large-scale instruction data are typically collected via web scraping or weak supervision, which inevitably introduces noisy or redundant samples, thereby hindering the capability~\cite{li2021improvedregularizationrobustnessfinetuning, szep2024practicalguidefinetuninglanguage,yang2024clip}. While manually constructing high-quality SFT datasets is ideal, it is prohibitively expensive at scale. As a practical alternative, recent research focuses on data selection methods that identify high-quality subsets from existing SFT data~\cite{li2025scardataselectionstyle, xia2024lessselectinginfluentialdata, wang2023far}. Such approaches have shown that models trained on selected subsets can match or even surpass the performance of those trained on the full dataset, offering a promising way to maintain model quality while significantly reducing computational costs.

Recent advances, such as reward-model-based filtering~\cite{xu2025betterreasoningdataenhancing} and gradient-based pruning~\cite{xia2024lessselectinginfluentialdata}, have pushed this direction forward.
Although achieving promising results, many of these methods incur huge computational overhead~\cite{xie2023dataselectionlanguagemodels} or rely on task-specific validation sets~\cite{xia2024lessselectinginfluentialdata}, which may limit their scalability across diverse domains. 
Furthermore, recent studies have found that models fine-tuned on such data are prone to domain overfitting, along with degraded generalization across domains~\cite{li2021improvedregularizationrobustnessfinetuning, szep2024practicalguidefinetuninglanguage}, which poses a unique challenge for ad-hoc instruction fine-tuning beyond the general SFT.
Therefore, these issues underscore the pressing need for a more principled, data-efficient view to scale LLM fine-tuning more cost-effectively and robustly, with the potential to reduce compute overhead and amplify impact in real-world deployments substantially.
This raises a central question:
\textit{How can we select the most representative and generalizable samples from large-scale instruction datasets to enable efficient and robust LLM fine-tuning?}
\begin{figure}[t]
\begin{center}
\includegraphics[width=0.75\linewidth]{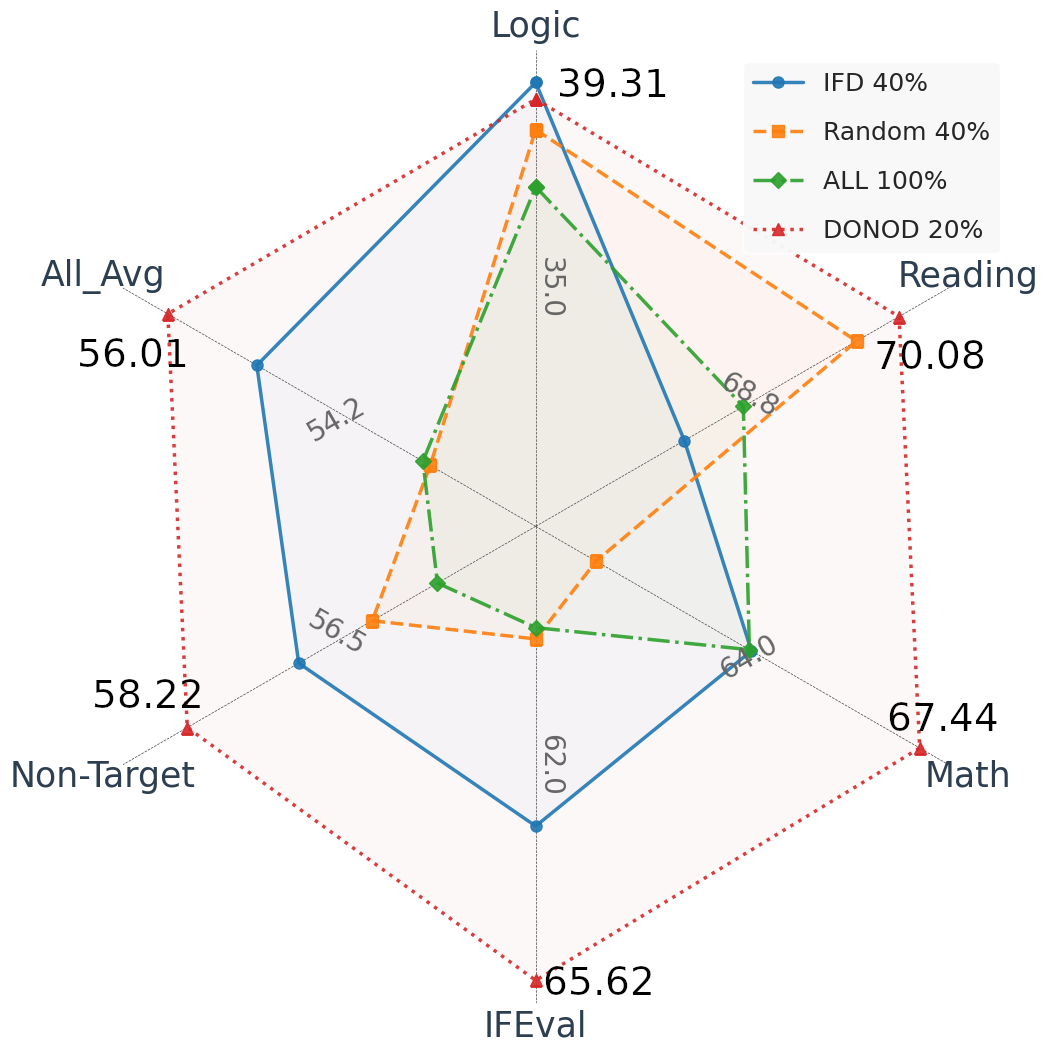}
\caption{Performance of our proposed DONOD across various evaluation benchmarks, consistently outperforming other baselines, e.g., all data, despite using only 20\% of the training data. These results highlight the potential of principled data pruning for efficient and robust LLM fine-tuning.
}
\label{fig:enter-mainresults}
\end{center}
\end{figure}

To address the issues above, we propose DONOD, a model-intrinsic dataset pruning method that selects a significantly reduced subset of the most representative samples.
Specifically, DONOD introduces two complementary metrics derived from the model's training dynamics: Delta of Norm (DON) and Norm of Delta (NOD).
The DON metric captures the cumulative shift in parameter magnitudes throughout training, serving as a proxy for generalization, which encourages the selection of samples that lead to stable norm evolution and contribute to smooth convergence.
Meanwhile, NOD measures the instability induced by individual samples by quantifying sample-specific weight fluctuations, thereby identifying noisy or unlearnable data.
To reconcile these dual objectives, maximizing generalization via DON and minimizing ambiguous updates via NOD, we employ the Technique for Order of Preference by Similarity to the Ideal Solution (TOPSIS) algorithm~\cite{Hwang1981,chakraborty2022topsis}, which allows principled ranking of samples based on proximity to the ideal pruning criterion. 
Importantly, DONOD requires no auxiliary models, domain-specific heuristics, or validation sets, relying purely on intrinsic training signals, thus ensuring computational efficiency and broad applicability.
As a result, DONOD enables efficient instruction tuning while preserving or even enhancing generalization performance with significantly fewer training examples.

Experiment results across diverse downstream instruction-tuning benchmarks and LLM architectures demonstrate that DONOD consistently outperforms state-of-the-art baselines, achieving superior performance with substantially fewer samples while maintaining dataset-agnostic flexibility.
Notably, compared with the full-data SFT setting, DONOD improves target-domain accuracy by 14.90\% and cross-domain accuracy by 5.67\% when using only 30\% of the data. 
Moreover, we demonstrate strong cross-architecture generalization, where data selected by lighter-weight models (e.g., LLaMA 3.1-8B~\cite{grattafiori2024llama3herdmodels}) can generalize well across larger or different architectures, such as LLaMA 2-13B~\cite{touvron2023llama2openfoundation} and Qwen 2.5-7B~\cite{qwen2.5}, etc.
Additionally, we further validate the robustness of DONOD in more complex and realistic noisy settings, highlighting its practical significance.
We summarize our contribution as follows: 
\begin{itemize}
    \item We propose DONOD, a lightweight, model-intrinsic data selection framework for accelerating the fine-tuning of LLMs.
    \item We introduce two complementary metrics, DON and NOD, where DON captures the generalization impact of samples and NOD identifies noisy or unlearnable samples.
    \item Experimental results validate that our method outperforms existing methods in terms of cross-domain and cross-architecture generalization performance, e.g., achieving lossless training acceleration with only 20\% of the data, highlighting the scalability and robustness of our approach.
\end{itemize}


\section{Related Work}
Dataset pruning for supervised fine-tuning on LLMs critically impacts model performance. Traditional methods often rely on external models as quality judges (\cite{du2023modsmodelorienteddataselection}, \cite{chen2024alpagasustrainingbetteralpaca}) or employ reward models to identify high-quality data~\cite{yang2025rl}. However, this dependence on auxiliary models, whether trained from scratch or repurposed, incurs significant computational costs and limits scalability. Recent studies (\cite{li2025preferenceleakagecontaminationproblem}) further question the effectiveness of this paradigm.

Alternative approaches focus on intrinsic data metrics. For instance, \cite{cao2024instructionmininginstructiondata} proposes evaluating data quality through features like length, naturalness, and coherence. However, the field lacks consensus on universal metrics: while \cite{chen2023maybe05dataneeded} emphasizes diversity, \cite{liu2024makesgooddataalignment} argues for prioritizing complex or challenging samples. This ambiguity motivates the third category—model-intrinsic methods—which leverage the model’s training dynamics to bypass explicit metric definitions. As noted by \cite{jiang2019acceleratingdeeplearningfocusing,yang2025dynamic}, the model’s response to data inherently signals its utility for learning, enabling automated dataset pruning.
\begin{figure*}[]
\centering
\includegraphics[width=.75\textwidth]{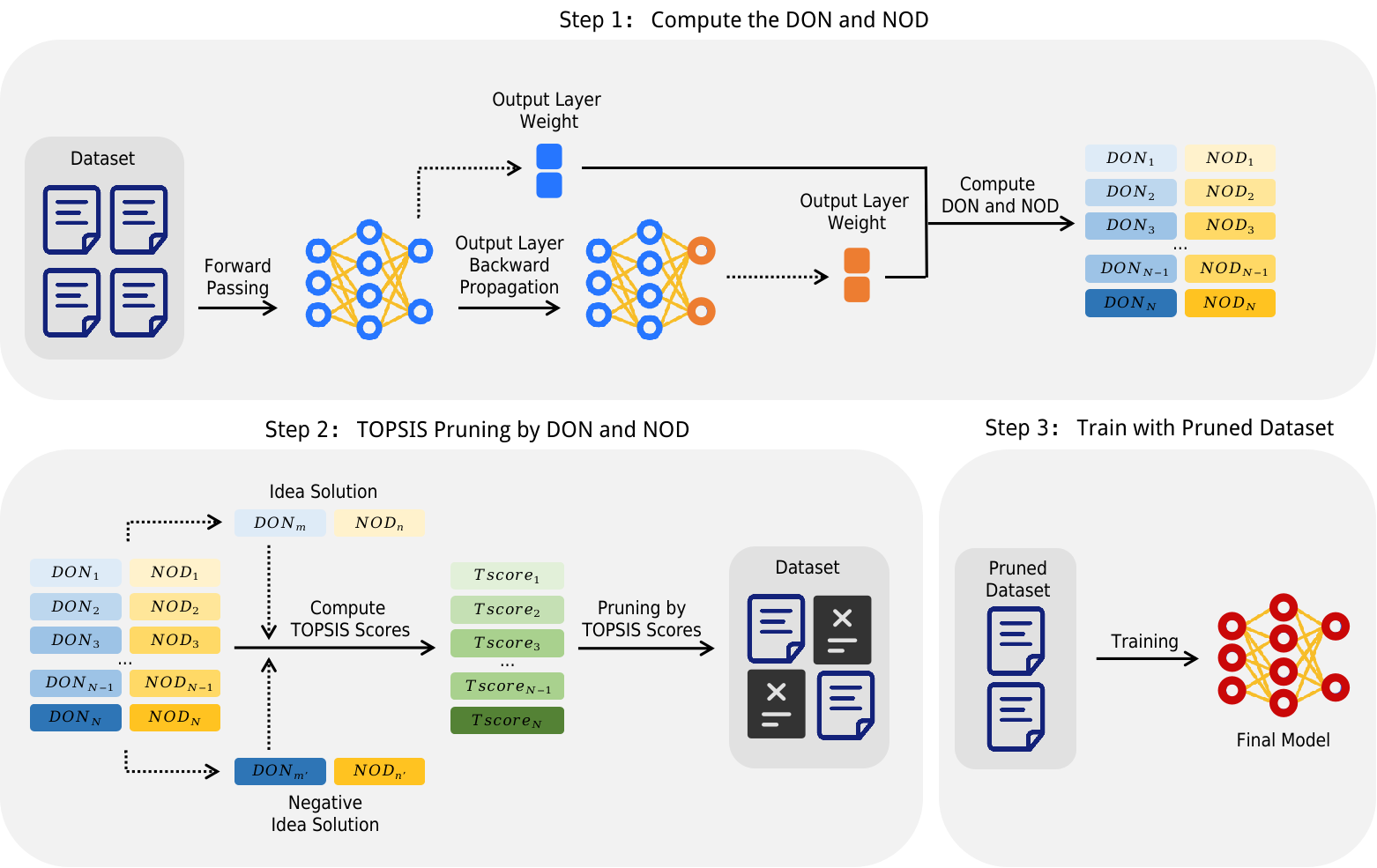}
\caption{Overview of our proposed DONOD, which follows a lightweight, three-step pipeline: (1) Compute DON and NOD metrics for each sample, (2) Apply TOPSIS pruning to select representative data and filter harmful/low-quality data, and (3) Train the model on the pruned dataset.}
\vspace{-.3cm}
\end{figure*}

Model-intrinsic pruning methods branch out based on various scenarios. The first category assumes access to a target data distribution, often via validation or development sets. For example, \cite{mindermann2022prioritizedtrainingpointslearnable} approximates loss differences between holdout and training sets, while \cite{xia2024lessselectinginfluentialdata} uses gradient similarity between validation and training data. These methods falter when target distributions are ambiguous or undefined, which is common in scenarios aiming to enhance broad capabilities rather than optimize for specific benchmarks.

The second category eliminates reliance on target distributions. Works like \cite{wang2024greats}, \cite{jiang2019acceleratingdeeplearningfocusing}, \cite{loshchilov2016onlinebatchselectionfaster}, and \cite{li2024quantityqualityboostingllm} employ loss or perplexity thresholds, assuming high-loss samples are valuable learning challenges. However, this assumption proves brittle for noisy or mislabeled data~\cite{yang2024clip}, where high loss reflects annotation errors rather than learnable patterns. Furthermore, challenging samples may exceed the model’s current capacity, rendering them unproductive for training.

Both categories neglect cross-domain generalization. Methods targeting specific distributions risk catastrophic forgetting, where performance gains on target tasks degrade generalizability. Conversely, loss-based selection exacerbates this by prioritizing samples that induce significant weight updates, destabilizing pre-trained knowledge.

To address these limitations, we propose DONOD. DON functions as a proxy of generalization, while NOD recognizes that the sample causes significant instability in the model weight. Integrated via the TOPSIS, DONOD filters noisy, unlearnable, and generalization-harming samples without auxiliary models or predefined targets. 

\section{The Proposed Method} \label{method}
\subsection{Overview}
Traditional SFT methods are highly data-dependent, and using a narrowly focused dataset can lead to catastrophic forgetting, resulting in performance degradation in non-target domains during specialization. For instance, fine-tuning on mathematical tasks may significantly impair the model’s ability in reading comprehension. While mixing diverse data can mitigate this, it introduces additional training costs and issues of domain proportion balance. Furthermore, the quality of individual samples significantly impacts fine-tuning efficacy, yet identifying high-quality data remains an open problem.

To this end, DONOD introduces a data-efficient alternative to select a reduced subset of representative samples and achieve SFT acceleration.
Specifically, DONOD consists of three core components: 1) DON and NOD metrics based on the Frobenius norm are used to estimate the samples' impact on model weight update.
2) TOPSIS is a multi-objective decision mechanism that balances task-specific gain and cross-domain generalization, ensuring that the selected subset preserves both in-domain effectiveness and robustness to distribution shifts. 3) By approximating the full model behavior through changes in the output layer, DONOD enables lightweight and scalable selection without the need to backpropagate through the entire model.
These components form an efficient data selection framework that supports accurate, low-cost subset selection, enabling fine-tuning with significantly fewer samples while maintaining or even improving model performance.

\subsection{DON and NOD Metrics} \label{sec:outputlayer}
Let $D$ denote an ad-hoc dataset for instruction fine-tuning. For a specific data sample $D_i \in D$, let $\{W^l\}_l^L$ represent the weight of the model before fine-tuning and  $\{W'^l\}_l^L$ the weight matrix after fine-tuning on $D_i$. We employ the DON and NOD to quantify this change. Specifically, the DON is defined as:
\begin{equation}
\begin{aligned}
\text{DON} &= \sum_{l=1}^L \left( \| W^l \|_F - \| W'^l \|_F \right) \\
&= \sum_{l=1}^L \left( \sqrt{\sum_{i=1}^{m_l} \sum_{j=1}^{n_l} |w^l_{i,j}|^2} - \sqrt{\sum_{i=1}^{m_l} \sum_{j=1}^{n_l} |w'^l_{i,j}|^2} \right),
\label{eq:don}
\end{aligned}
\end{equation}
where $m_l$ and $n_l$ are the dimensions of the weight matrix $W^l$ of layer $l$, $\| \cdot \|_F$ denotes the Frobenius norm. 
Here, we adopt the Frobenius norm due to its ability to capture fine-grained structural changes across all weight elements while maintaining computational efficiency, making it a suitable and scalable proxy for quantifying sample-level influence in large-scale models.
Thus, DON captures the cumulative shift in the model’s weight magnitude.
From a generalization perspective, a positive DON suggests that a sample reduces the model’s Frobenius norm, which is associated with lower complexity and better generalization~\cite{NIPS1996_fb2fcd53, yin2020rademachercomplexityadversariallyrobust, 10.5555/2621980}. 
Meanwhile, the NOD is defined as:
\begin{equation}
\begin{aligned}
\text{NOD} = \sum_{l=1}^L \| W^l - W’^l \|_F =  \sum_{l=1}^L \sqrt{ \sum_{i=1}^{m_l} \sum_{j=1}^{n_l} \|w^l_{i,j} - w’^l_{i,j}\|^2 }, \label{eq:nod}
\end{aligned}
\end{equation}
which measures the direct geometric displacement of weights in parameter space at the current step.
Thus, in the context of SFT, NOD quantifies how drastically a single sample perturbs the parameter space. 
These metrics are complementary: DON reflects the overall scaling of weights, whereas NOD reflects the sensitivity of model weight on a single sample.


While the Frobenius norm can be applied to whole model weights, prior works~\cite{nadipalli2025layerwiseevolutionrepresentationsfinetuned,rosati2024representation} show that fine-tuning primarily affects later layers, with the output layer acting as a bottleneck for domain adaptation.
Therefore, we estimate the sample influence using the weights of the last layer, which brings two benefits: 1) Computational Efficiency: The output layer is typically smaller than the hidden layers, reducing the computational cost of computing norms across iterations, 2) Interpretability: Output layer updates correlate more directly with task performance, avoiding the entangled representations of deeper layers. 

In this way, we derive a simplified version of the computation of \eqref{eq:don} and \eqref{eq:nod}:
\begin{equation} \begin{split} \text{DON} &= \| W^L \|_F - \| W'^L \|_F \\ &= \sqrt{\sum_{i=1}^{m_L} \sum_{j=1}^{n_L} |w^L_{i,j}|^2} - \sqrt{\sum_{i=1}^{m_L} \sum_{j=1}^{n_L} |w'^L_{i,j}|^2} ,\end{split} \end{equation}
where \( m_L \) and \( n_L \) are the dimensions of the output layer.
\begin{equation}  
\text{NOD} = \| W^L - W'^L \|_F = \sqrt{\sum_{i=1}^{m_L} \sum_{j=1}^{n_L} |w^L_{i,j} - w'^L_{i,j}|^2}.
\end{equation}


  

\subsection{Integration of TOPSIS}
TOPSIS is a multi-criteria decision analysis (MCDA) method that ranks alternatives by their relative closeness to an ideal solution. In DONOD, TOPSIS is employed to resolve the inherent tension between the two metrics—DON and NOD—by identifying samples that simultaneously maximize DON (to enhance generalization) and minimize NOD (to avoid noise).
After computing the DON and NOD for each sample, we integrate TOPSIS to rank the data points. 

TOPSIS inherently balances conflicting objectives, i.e., maximizing DON and minimizing NOD, by leveraging geometric distance in the normalized metric space. Moreover, normalization mitigates the impact of differing metric magnitudes, ensuring neither DON nor NOD dominates the ranking. It avoids subjective weight assignment (unlike weighted sum) and provides a total ordering of samples (unlike Pareto optimality), which is critical for deterministic pruning decisions. Thus, we choose TOPSIS in our framework.

\paragraph{Computational Complexity}
\label{sec:complexity}
The algorithm's computational complexity consists of: (1) per-sample DON and NOD computation and (2) TOPSIS-based sample selection.

In the first phase, we perform a forward pass (\(O(P)\) time per sample, where \(P\) is the model's parameter count), a backward pass restricted to the output layer (\(O(O)\) time, \(O\) being the output layer's parameters), and compute Frobenius norms for weight updates (\(O(O)\)). Since \(O \subset P\), the per-sample cost simplifies to \(O(P)\), resulting in a total training complexity of \(O(N \cdot P)\), where \(N\) is the number of training samples.

The second phase involves normalizing DON/NOD metrics (\(O(N)\)), computing distances to ideal and negative-ideal solutions (\(O(N)\)), and ranking samples via TOPSIS scores, dominated by an \(O(N \log N)\) sorting step. Thus, the selection process has a total complexity of \(O(N \log N)\).

Combining both phases, the dominant term is \(\mathcal{O}(N \cdot P)\), as \(P \gg N \log N\) in modern neural networks (e.g., \(P \sim 10^6\)–\(10^{12}\) parameters). The overall time complexity simplifies to $\boxed{\mathcal{O}(N \cdot P)}$.
 For storage, only \(O(N)\) space is required to store per-sample metrics, as no intermediate model states need to be retained. Therefore, the runtime of DONOD can be approximated by the model's inference speed. In our experiments using Llama-3.1-8B-Instruct, processing the SAT Math COT dataset took approximately 18 minutes of wall-clock time on a single A100 80GB GPU, with negligible storage requirements.

\section{Experiment}
\subsection{Experiment Setup}
\paragraph{Evaluation Benchmarks and Training Datasets}
Following \cite{ma2025korbench}, we construct our evaluation benchmark based on AGIEval~\cite{zhong-etal-2024-agieval} and IFEval~\cite{zhou2023instructionfollowingevaluationlargelanguage}. 
This benchmark is designed to be comprehensive and domain-orthogonal, assessing abilities in logical reasoning, mathematics, reading comprehension, and instruction following.
By incorporating diverse datasets, the benchmark reflects real-world ad-hoc SFT scenarios, where the objective is to strengthen targeted model abilities rather than optimize for narrow or unrepresentative benchmarks.
\begin{table*}[]
\centering
\scalebox{0.85}{%
\begin{tabular}{
  >{\centering\arraybackslash}m{3.5cm} 
  l 
  *{6}{S[table-format=2.2]}
}
\toprule
\textbf{Dataset} & \textbf{Method} & 
\multicolumn{1}{c}{\textbf{Logic$\uparrow$}} & 
\multicolumn{1}{c}{\textbf{Reading$\uparrow$}} & 
\multicolumn{1}{c}{\textbf{Math$\uparrow$}} & 
\multicolumn{1}{c}{\textbf{IFEval$\uparrow$}} & 
\multicolumn{1}{c}{\textbf{Non-Target$\uparrow$}} & 
\multicolumn{1}{c}{\textbf{All Avg$\uparrow$}} \\
\midrule

\multirow{5}{*}{\textbf{LogiQA Train}} & LESS 5\%       & 22.62 & \textbf{25.19} & 21.47 & 64.33 & 36.99 & 27.91 \\
                                 & Random 5\%     & 20.45 & 24.27 & 24.06 & 46.95 & 31.96 & 26.35 \\
                                 & ALL 100\%      & \textbf{22.98} & 24.02 & 22.26 & 17.74 & 21.34 & 22.77 \\
                                 & DONOD 5\%      & 20.91 & 24.27 & \textbf{24.22} & \textbf{68.95} & \textbf{39.15} & \textbf{28.86} \\
\midrule

\multirow{3}{*}{\textbf{GSM8K}} & ALL (100\%)     & 33.43 & 70.74 & \textbf{47.99} & 59.70 & 48.96 & \textbf{48.74} \\
                                & LESS 5\%        & 34.74 & 71.85 & 25.78 & 60.63 & 50.46 & 44.97 \\
                                & DONOD 30\%      & \textbf{37.16} & \textbf{71.92} & 33.69 & \textbf{71.16} & \textbf{52.74} & 48.51 \\
\bottomrule
\end{tabular}%
}
\caption{Experimental results with domain-specific averages on LogiQA Train and GSM8K. The \textbf{Non-Target} column shows average performance excluding logic reasoning or Math (target domain).\label{tab:results_logiq_gsm8k}}
\end{table*}

To comprehensively evaluate DONOD's effectiveness across diverse domains, tasks, and data conditions, we select data that span a wide range of settings, including variations in domain (e.g., mathematics, logical reasoning, instruction following), task format (e.g., chain-of-thought, multiple-choice, fine-grained evaluation), and the presence or absence of validation sets. This design enables robust benchmarking under realistic and varied constraints.
Specifically, we assess DONOD across the following settings:
\begin{itemize}
    \item \textbf{SAT Math Chain-of-Thought (COT)} \cite{davidson2023satmath} (math, COT, no validation set).
    \item \textbf{LogiQA-Train} \cite{liu2020logiqachallengedatasetmachine} (logical reasoning, multiple-choice, with validation set).
    \item \textbf{IFEval-like Data} \cite{xu2024magpiealignmentdatasynthesis} (instruction-following, general, no validation set).
    \item \textbf{GSM8K} \cite{cobbe2021gsm8k} (math, COT, biased distribution vs. SAT Math and Aqua-RAT).
\end{itemize}

\begin{table*}[]
\centering

\scalebox{0.85}{%
\begin{tabular}{>{\centering\arraybackslash}m{3.5cm} l *{6}{S[table-format=2.2]}}
\toprule
\textbf{Dataset} & \textbf{Method} & 
\multicolumn{1}{c}{\textbf{Logic$\uparrow$}} & 
\multicolumn{1}{c}{\textbf{Reading$\uparrow$}} & 
\multicolumn{1}{c}{\textbf{Math$\uparrow$}} & 
\multicolumn{1}{c}{\textbf{IFEval$\uparrow$}} & 
\multicolumn{1}{c}{\textbf{Non-Target$\uparrow$}} & 
\multicolumn{1}{c}{\textbf{All Avg$\uparrow$}} \\
\midrule
\multirow{5}{*}{\textbf{SAT Math COT}} & IFD 40\%        & 39.31 & 68.26 & 64.18 & 63.03 & 56.87 & 55.04 \\
& ALL 100\%       & 37.12 & 68.76 & 64.14 & 59.70 & 55.19 & 53.23 \\
& Random 40       & 38.32 & 69.72 & 61.17 & 59.89 & 55.98 & 53.15 \\
& DONOD 30\%      & \textbf{39.98} & 67.62 & \textbf{73.70} & 63.40 & 57.00 & \textbf{56.25} \\
& DONOD 20\%      & 38.96 & \textbf{70.08} & 67.44 & \textbf{65.62} & \textbf{58.22} & 56.01 \\
\midrule
\multirow{3}{*}{\textbf{IFEval-Like Data}} & ALL (100\%)     & 36.06 & 64.92 & 44.84 & 64.33 & 48.60 & 48.68 \\
& IFD 40\%        & 34.06 & \textbf{66.31} & 34.04 & \textbf{71.90} & 44.80 & 46.55 \\
& DONOD 30\%      & \textbf{36.14} & 57.82 & \textbf{66.47} & 46.77 & \textbf{53.47} & \textbf{49.01} \\
\bottomrule
\end{tabular}%
}
\caption{Experimental results with domain-specific averages on SAT Math COT and IFEval-Like Data. The \textbf{Non-Target} column shows average performance excluding mathematical reasoning or IFEval (target domain), revealing how methods generalize to other abilities. The \textbf{All Avg} column shows the average of all tasks in the benchmark. All values are percentages.\label{tab:sat_IFEval}}
\end{table*}

\paragraph{Models and Experiment Settings}
We evaluate DONOD on a diverse set of instruction-tuned models to assess its generalizability across architectures and fine-tuning paradigms. Specifically, we consider: (1) LLaMA-3.2-3B-Instruct~\cite{meta2024llama32blog}, a lightweight model optimized for instruction-following tasks; (2) LLaMA-3.1-8B-Instruct~\cite{grattafiori2024llama3herdmodels}, a mid-sized model widely used in recent instruction-tuning studies; (3) LLaMA-2-13B-Chat~\cite{touvron2023llama2openfoundation}, a larger model trained with conversational objectives; and (4) Qwen 2.5-7B-Instruct~\cite{qwen2.5}, a model from a distinct architecture family, differing in tokenizer, training data, and parameterization.
This ensures a comprehensive evaluation of DONOD under varied model designs and training strategies, i.e., cross-architecture generalization and supervised fine-tuning performance.
As described in Section~\ref{sec:outputlayer}, we focus on the output layer of the Llama-3.1-8B-Instruct model for our experiment.

\subsection{Main Results} \label{mainresults}

\begin{table}[]
\centering
\small

\scalebox{0.8}{%
\begin{tabular}{lcccccc}
\toprule
\textbf{Method} & 
\multicolumn{1}{c}{\bf \makecell{Logic\\(↑)}} & 
\multicolumn{1}{c}{\bf \makecell{Reading\\(↑)}} & 
\multicolumn{1}{c}{\bf \makecell{Math\\(↑)}} & 
\multicolumn{1}{c}{\bf \makecell{IFEval\\(↑)}} & 
\multicolumn{1}{c}{\bf \makecell{Non-Target\\(↑)}} & 
\multicolumn{1}{c}{\bf \makecell{All\\Avg (↑)}} \\
\midrule
ALL (100\%)      & 43.62 & 73.99 & 74.57 & 56.01 & 57.87 & 58.90 \\
DONOD 20\%       & 41.95 & 68.28 & 79.86 & 58.23 & 56.15 & 59.09 \\
DONOD 30\%       & 42.57 & 71.15 & 77.07 & 61.74 & 58.48 & 59.33 \\
\bottomrule
\end{tabular}}
\caption{Cross-architecture generalization of selecting data with Llama-3.1-8B-Instruct and fine-tuning on Qwen-2.5-7B-Instruct.}\label{tab:cross-arc-qwen}
\end{table}

\paragraph{Comparison with State-of-the-arts}  
As shown in Table~\ref{tab:results_logiq_gsm8k} and Table~\ref{tab:sat_IFEval}, DONOD consistently outperforms others while using significantly less data across nearly all benchmarks, e.g., 20-30\%, and requiring no development set. It achieves state-of-the-art performance in core reasoning tasks such as math and logic, outperforming full-data baselines in both target-domain accuracy and cross-domain generalization.

Compared to IFD and LESS, DONOD demonstrates better balance between accuracy and efficiency, achieving higher overall and non-target averages under constrained data budgets. Notably, it exhibits strong cross-task and cross-domain transferability, without relying on task-specific tuning or heuristics, and remains competitive even in challenging settings, such as reading comprehension.

These results highlight DONOD’s ability to retain or exceed full-data performance while substantially reducing training cost, which offers a robust and scalable solution for data-efficient LLM fine-tuning.

\begin{table}[]
\centering
\small

\scalebox{0.8}{%
\begin{tabular}{lcccccc}
\toprule
\textbf{Method} & 
\multicolumn{1}{c}{\bf \makecell{Logic\\(↑)}} & 
\multicolumn{1}{c}{\bf \makecell{Reading\\(↑)}} & 
\multicolumn{1}{c}{\bf \makecell{Math\\(↑)}} & 
\multicolumn{1}{c}{\bf \makecell{IFEval\\(↑)}} & 
\multicolumn{1}{c}{\bf \makecell{Non-Target\\(↑)}} & 
\multicolumn{1}{c}{\bf \makecell{All\\Avg (↑)}} \\
\midrule
ALL (100\%)      & 30.57 & 54.68 & 23.38 & 29.94 & 38.39 & 35.28 \\
DONOD 20\%       & 31.52 & 55.58 & 26.89 & 29.39 & 38.83 & 36.57 \\
DONOD 30\%       & 31.30 & 54.61 & 23.39 & 29.57 & 38.49 & 35.53 \\
\bottomrule
\end{tabular}}
\caption{Cross-architecture generalization of selecting data with Llama-3.1-8B-Instruct and fine-tuning on Llama-2-13b-chat.}\label{tab:cross-arc-llama}
\end{table}

\begin{table*}[t]  
\centering

\scalebox{0.9}{%
\begin{tabular}{
  >{\centering\arraybackslash}m{3.5cm} 
  l 
  *{6}{S[table-format=2.2]}
}
\toprule
\textbf{Dataset} & \textbf{Method} & 
\multicolumn{1}{c}{\textbf{Logic$\uparrow$}} & 
\multicolumn{1}{c}{\textbf{Reading$\uparrow$}} & 
\multicolumn{1}{c}{\textbf{Math$\uparrow$}} & 
\multicolumn{1}{c}{\textbf{IFEval$\uparrow$}} & 
\multicolumn{1}{c}{\textbf{Non-Target$\uparrow$}} & 
\multicolumn{1}{c}{\textbf{All Avg$\uparrow$}} \\
\midrule

\multirow{5}{*}{\textbf{Main Results}} & DON               & 36.19 & 68.12 & 58.15 & \textbf{66.17} & 56.83 & 52.38 \\
                                       & NOD               & \textbf{40.23} & 67.12 & 65.71 & 58.04 & 55.13 & 54.39 \\
                                       & Weighted Sum        & 38.08 & 69.15 & 63.74 & 53.79 & 53.67 & 53.54 \\
                                       & Pareto Optimization & 36.85 & 70.02 & 57.82 & 65.25 & 57.37 & 53.07 \\
                                       & DONOD             & 38.96 & \textbf{70.08} & \textbf{67.44} & 65.62 & \textbf{58.22} & \textbf{56.01} \\
\bottomrule
\end{tabular}%
}
\caption{Experimental results with domain-specific averages. The \textbf{Non-Target} column shows average performance (\%) excluding mathematical reasoning (target domain), revealing how methods generalize to other abilities.}
\label{tab:results}
\end{table*}

\subsection{Cross-Architecture Generalization}
To evaluate cross-architecture generalization, we select data using Llama-3.1-8B-Instruct and fine-tune on both Llama-2-13b-chat and a structurally different model, Qwen-2.5-7B-Instruct. 
As shown in Table~\ref{tab:cross-arc-qwen} and Table~\ref{tab:cross-arc-llama}, DONOD-selected subsets (20\% and 30\%) not only retain but surpass the full-data baseline in overall performance.
These results suggest that despite differences in model size and structure, language models exhibit consistent rankings in perceiving instruction difficulty.
In~\cite{li2024superfilteringweaktostrongdatafiltering}, this consistency is demonstrated through metrics like perplexity and Instruction-Following Difficulty scores, which show strong rank correlations across models of different sizes. As a result, smaller models like GPT-2 can effectively filter instruction data for much larger models, such as LLaMA2-7B or GPT-4.

DONOD leverages this same principle by selecting data based on intrinsic parameter-level signals. These signals identify samples that are universally useful across architectures. In these experiments, we observe that pruning data using DONOD on a smaller model (e.g., Llama-3.1-8B) leads to better performance when fine-tuning larger or different models (e.g., Llama-2-13B and Qwen-2.5-7B), echoing the weak-to-strong transfer effect.
These results demonstrate strong transferability of selected samples, highlighting DONOD’s practical utility in real-world settings.

\subsection{Robustness in Identifying Noise}
To evaluate DONOD’s robustness in identifying noisy data, we conduct a controlled experiment on the SAT Math CoT dataset using the LLaMA 3.1 model. 
We start with the top 20\% of samples originally selected by DONOD.

To simulate real-world data imperfections, we introduce controlled noise by randomly masking words in the labels of these clean samples, which mimics subtle corruption or annotation errors.
These perturbed samples are then reintegrated into the full dataset, creating a new training pool containing embedded noisy instances.
We reapply DONOD to this dataset to select a new top 20\% subset based on the updated DON and NOD values.
We assess sensitivity to noise by measuring the overlap between the original and newly selected top 20\%. The result shows a drop to only 38.7\% overlap, indicating that DONOD successfully identifies and filters out many of the newly corrupted samples.

This experiment highlights DONOD’s strong responsiveness to fine-grained label corruption and its ability to dynamically adapt selection criteria. Such robustness is critical in real-world scenarios where large-scale instruction data, often web-scraped or weakly supervised, is prone to noise and inconsistency.

\subsection{Ablation Study}
We conduct experiments using LLaMA-3.1-8B-Instruct on the SAT Math CoT dataset to analyze the effect of different components in DONOD. 
 As shown in Table ~\ref{tab:results}, we evaluate four configurations: (1) DON-only ranking, (2) NOD-only ranking, (3) DON + NOD combinations without TOPSIS (Weighted Sum and Pareto Front), and (4) the full method (DON + NOD + TOPSIS). Notably, we omit ablations combining TOPSIS with only one metric (e.g., DON + TOPSIS) because TOPSIS inherently requires multiple criteria to resolve conflicting rankings. 

\paragraph{Impact of Individual Metrics}
When using NOD alone, we observe stronger performance in the target domain (e.g., Math: 65.71\%) compared to DON (58.15\%), as NOD emphasizes samples that lead to significant, localized parameter updates, favoring task-specific learning. However, this comes at the cost of reduced generalization to non-target domains, suggesting overfitting to task-specific patterns.

In contrast, DON alone yields better cross-domain generalization by favoring samples that contribute to stable and smooth parameter updates, which better preserve generalizable knowledge. However, this approach underperforms in task-specific adaptation, reflecting a trade-off between stability and specificity.

\paragraph{Combination Strategies}
Simply combining DON and NOD via weighted sum or Pareto Front fails to effectively reconcile the competing goals of DON and NOD. 
The weighted sum approach marginally improves Reading but degrades Math and Non-Target compared to using DON or NOD individually. 
Similarly, the Pareto Front strategy emphasizes cross-domain stability but sacrifices gains in the target domain. These results underscore the difficulty of balancing conflicting criteria without a principled integration mechanism.

\paragraph{Full Method (DONOD)}
The proposed DONOD achieves the best trade-off between domain-specific performance and cross-domain generalization.
By ranking samples based on their proximity to the ideal combination of DON and NOD, TOPSIS effectively resolves conflicts between the two objectives. This leads to improvements in both target-domain accuracy and cross-domain robustness. The overall average performance also improves significantly, validating the effectiveness and importance of using TOPSIS as a principled strategy for multi-objective optimization.




\subsection{More Analytical Results}
\begin{figure}[t]
    \centering
        \includegraphics[width=0.5\linewidth]{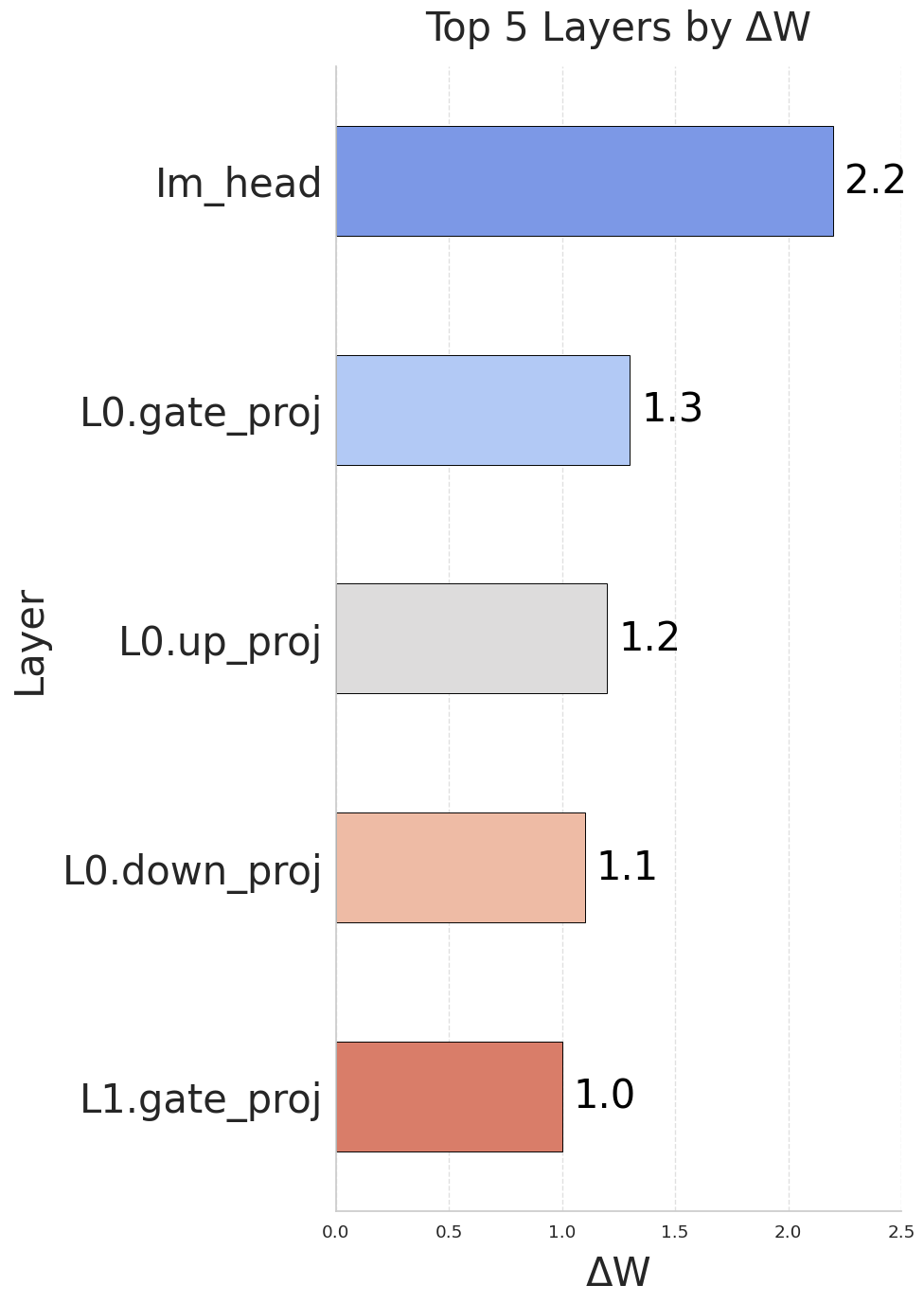}
        \caption{Rank of the top 5 layers by Frobenius norm delta after fine-tuning Llama-2-13B-Chat on SAT Math COT.}
        \label{fig:ranking}
\end{figure}
\begin{figure}[t]
\begin{center}
\includegraphics[width=0.95\linewidth]{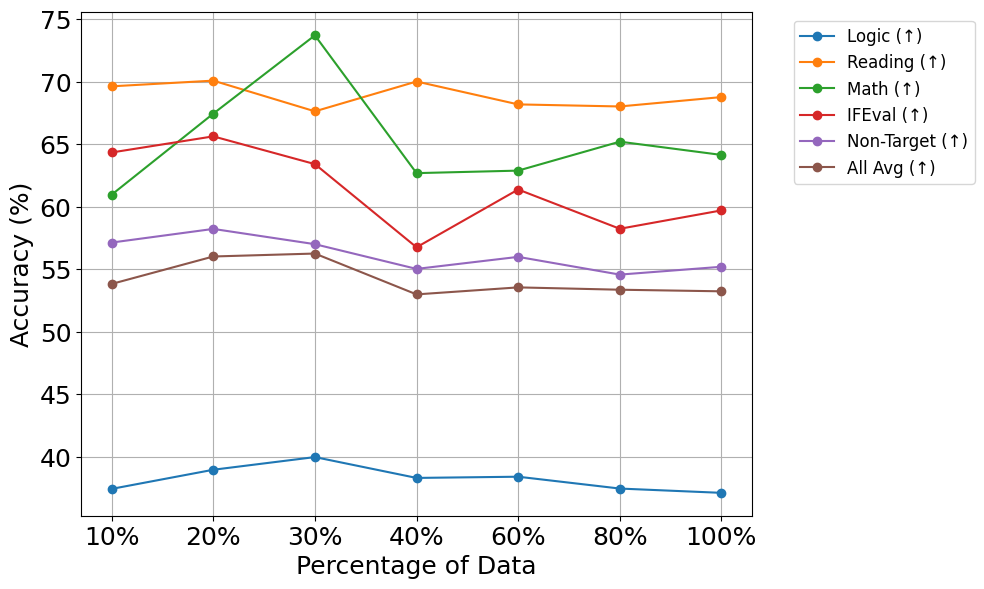}
\caption{Stability of DONOD across data proportions. }
\label{fig:dynamics}
\end{center}
\end{figure}



\paragraph{Validation of Output Layer Focus}
\label{sec:exp_outputlayer}
To further justify our design choice in Section~\ref{sec:outputlayer}, we analyze the sensitivity of weight changes across layers. Ranking the Frobenius norm delta across layers after fine-tuning Llama-2-13B-Chat on the SAT Math COT dataset, as shown in Figure~\ref{fig:ranking}, the output layer exhibits the largest delta, indicating its heightened responsiveness to task-specific supervision.
By restricting DON and NOD to this layer, we retain a representative signal of overall weight dynamics (closely aligned with the layer-wise average Frobenius norm) while significantly reducing computational overhead. 
In contrast to full backpropagation and computing $DON_{total}$ and ${NOD_{total}}$ across all layers for every data sample, our approach only requires backpropagation through the final layer, along with a constant time $DON$ and $NOD$ computation, yielding a highly efficient yet effective approximation.

\paragraph{Stability of DONOD Across Data Proportions}
To evaluate the stability of DONOD under varying data regimes, we analyze the performance when trained on the selected datasets (10\%–100\%). As shown in Figure~\ref{fig:dynamics}, DONOD demonstrates remarkable stability and efficiency, achieving competitive or better performance even with significantly reduced data.

Notably, with only 10\% of the data, DONOD surpasses the full-dataset baseline in Logic and Overall Average, highlighting its robustness against data scarcity. At a selection ratio of 20\%, our method achieves an optimal balance between data efficiency and model performance, achieving peak scores in Logic and Reading, while maintaining a strong overall average. 
This suggests that DONOD effectively identifies high-quality samples that maximize learning signals with minimal noise. 

\paragraph{Human Evaluation on Selected Datasets}
\begin{figure}[]
\begin{center}
\includegraphics[width=0.75\linewidth]{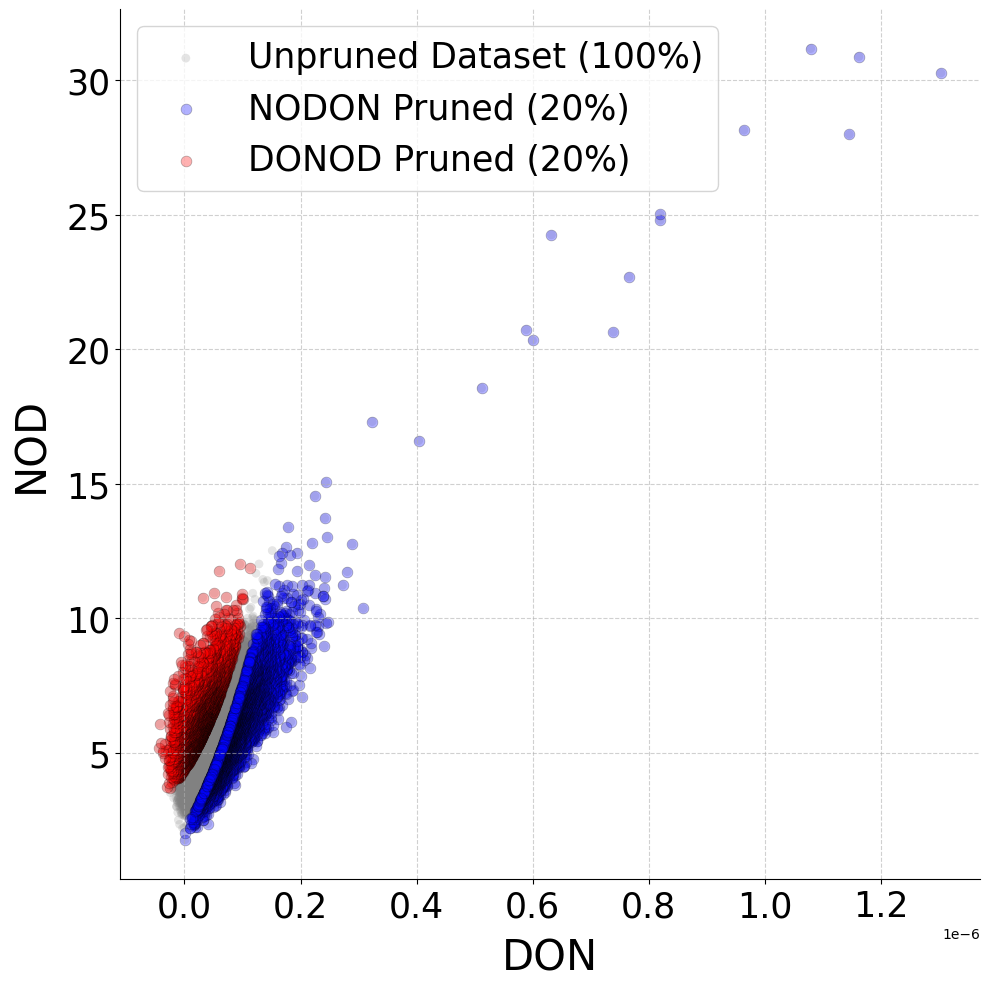}
\caption{Illustration of the distribution of the selected dataset with DONOD and NODON.
}
\label{pruned_distribution}
\end{center}
\end{figure}
To rigorously demonstrate the effectiveness of DONOD, a human evaluation is conducted. By reversing the selection process, where the top $N$ samples with the highest TOPSIS scores are retained, NODON generates a dataset containing the top samples that DONOD would filter out. As illustrated in Figure~\ref{pruned_distribution}, these samples tend to cluster on the right side of the NOD-DON plane.

To further analyze the data pruned by our method, we then examine the samples from the NODON-pruned SAT Math COT datasets. The evaluation reveals several common categories of pruned samples: 1) Overthinking in response to easy questions, 2) Incomplete or partial answers, 3) Incorrect answers due to reasoning errors, 4) Incorrect answers resulting from errors in the question itself, and 5) Overly complex or excessively difficult questions.
These findings suggest that DONOD effectively filters out unhelpful or misleading samples, thereby improving instruction data quality and enhancing the overall robustness and generalization of fine-tuned LLMs.

\section{Conclusion}
In this paper, we propose DONOD, a model-intrinsic dataset pruning framework to enhance LLM fine-tuning efficiency without sacrificing model performance. 
By leveraging weight dynamics, our method selects high-quality data and suppresses the selection of noisy or uninformative data via dual complementary metrics, DON and NOD.
These metrics are integrated via TOPSIS, enabling a principled trade-off between maximizing generalization and minimizing harmful updates.
Experiments show that DONOD can reduce training data volume by up to 70\% while outperforming standard supervised fine-tuning, achieving superior training acceleration. 
Notably, datasets selected by smaller models also generalize well when used to fine-tune other LLM architectures, underscoring the framework's scalability and practicality for real-world LLM pipelines.
We hope DONOD inspires further research on data selection for LLM training from a model-intrinsic perspective and believe our method will serve as a promising dataset optimization tool for the community, enabling enhanced data-centric LLM training pipelines.

\bibliography{aaai2026}
\clearpage
\clearpage
\appendix
\section{Appendix}

\subsection{Theoretical Foundations}\label{appendix:theoretical_found}
The generalization gap for a learning algorithm, for a specific function $f$, is the difference between its true (expected) risk $R_\mathcal{D}(f)$ over the data distribution $\mathcal{D}$ and its empirical risk $R_S(f)$ over a finite sample $S = \{ (x_1, y_1), \dots, (x_m, y_m) \}$. A smaller generalization gap means the model's performance on unseen data is closer to its performance on training data, i.e., the model generalizes well.

The Rademacher complexity $R_m(\mathcal{F})$ of a hypothesis class $\mathcal{F}$ measures its ability to fit random noise. It is formally defined (\cite{10.5555/944919.944944} and \cite{pmlr-v40-Neyshabur15}) as:
$$ 
R_m(\mathcal{F}) = \mathbb{E}_{\boldsymbol{\xi} \in \{\pm 1\}^m} \left[ \frac{1}{m} \sup_{f \in \mathcal{F}} \left| \sum_{i=1}^m \xi_i f(x_i) \right| \right] 
$$
Similarly by \cite{10.5555/944919.944944}, for a hypothesis class $\mathcal{F}$ of real-valued functions and a 1-Lipschitz loss function $\ell$, a standard generalization bound, if inputs are bounded and output is bounded, states that for any $f \in \mathcal{F}$, with probability at least $1-\delta$ over the random draw of the sample $S$:
$$ 
R_\mathcal{D}(f) \le R_S(f) + 2 R_m(\mathcal{F}) + C \sqrt{\frac{\log(1/\delta)}{m}} 
$$
where $C$ is a constant related to the range of the loss function, and for 1-Lipschitz losses, it simplifies to $2 R_m(\mathcal{F}) + \sqrt{\frac{\log(1/\delta)}{2m}}$.
This inequality implies that the generalization gap $R_\mathcal{D}(f) - R_S(f)$ is bounded by terms that include the Rademacher complexity of the hypothesis class $\mathcal{F}$. Therefore, to show that the generalization gap of $M'$ is smaller, we need to show that the Rademacher complexity of its corresponding hypothesis class is smaller.

\subsection{Theoretical Analysis}
\label{sec:introduction}
Here, based on our case, we simplify our network as a network with a fixed architecture ($d$ layers, width $H$) and RELU activations, and consider a hypothesis class of functions whose weights $w$ satisfy a bound on $\mu_{p,q}(w)$.
Specifically, we are interested in the Frobenius norm, which is $\mu_{2,2}(w)$.
Let $\mathcal{F}_{\mu} = N_{d,H,\sigma_{RELU}}^{\mu_{2,2} \le \mu}$ be the class of functions that can be realized by such a network where the overall $\ell_2$ norm of its weights is at most $\mu$.

For the model $M_{high}$ of weight  $W_{high}$, its actual Frobenius norm is $\mu_{M_{high}} = \mu_{2,2}(W_{high})$. The function $f_{W_{high}}$ computed by $M_{high}$ belongs to the hypothesis class $\mathcal{F}_{\mu_{M_{high}}}$.

Similarly, the network  $M_{low}$ has weights $W_{low}$, and its actual Frobenius norm is $\mu_{M_{low}} = \mu_{2,2}(W_{low})$. The function $f_{W_{low}}$ computed by $M_{low}$ belongs to the hypothesis class $\mathcal{F}_{\mu_{M_{low}}}$.
We are given $\mu_{M_{low}}>\mu_{M_{high}}$.

Here, we theoretically prove that models with high DON generalize better.

To resonate the use of DON, we first introduce two key mathematical tools, the generalization gap and Rademacher complexity, as shown in Appendix 1.1. 
Let $M$, $M_{high}$ and $M_{low}$ be three neural networks with identical architectures but distinct weight matrices $W$, $W_{low}$ and $W_{high}$, respectively. Here $M_{high}$ stands for a model with high DON, $M_{low}$ stands for a model with low DON, and $M$ is an arbitrary auxiliary model.

We show that the model with higher DON generalizes better, e.g., if 
\begin{equation}
\begin{split}
\text{DON}^{\text{high}} - \text{DON}^{\text{low}}
&= \sum_{l=1}^L \left( \| W^l \|_F - \| W^l_{\text{high}} \|_F \right) \\
&\quad - \sum_{l=1}^L \left( \| W^l \|_F - \| W^l_{\text{low}} \|_F \right) > 0
\end{split}
\label{eqation1}
\end{equation}

then $M_{high}$ generalize better.
    
Simplify \ref{eqation1}, we have:
\begin{equation} \label{eqation2}
    \sum_{l=1}^L ( \| W^l_{low} \|_F - \| W^l_{high} \|_F) > 0
\end{equation}
Using the same notation as \cite{pmlr-v40-Neyshabur15}, let $\mu_{2,2}(W_{low}) = \sum_{l=1}^L\| W^l_{low} \|_F$ and $\mu_{2,2}(W_{high}) = \sum_{l=1}^L\| W^l_{high} \|_F$, suppose the Frobenius norms of the weights satisfy  $\mu_{2,2}(W_{low}) > \mu_{2,2}(W_{high})$, we want to proof that that the generalization gap of $M_{high}$ is smaller than that of $M_{low}$.

\noindent \textbf{Proof.}
Firstly, we make necessary assumptions and the setup based on our practical conditions as detailed in Appendix 1.2. According to the  Corollary 2 of \cite{pmlr-v40-Neyshabur15}, for any $d \ge 1$, $1 \le p < \infty$, and $1 \le q \le p^* = p/(p-1)$ (where $1/p + 1/p^* = 1$), the Rademacher complexity of the class $N_{\mu_{2,2} \le \mu}^{d,H,\sigma_{RELU}}$ is bounded.

In our case, we are considering the Frobenius norm, so $p=2$ and $q=2$. This means $p^* = 2/(2-1) = 2$.
Since $q=2$ and $p^*=2$, the condition $q \le p^*$ ($2 \le 2$) is met. Therefore, we can apply the bound from Corollary 2:
$$ R_m(N_{\mu_{2,2} \le \mu}^{d,H,\sigma_{RELU}}) \le \left( \frac{2\mu}{\sqrt[2]{d}} \right)^d R^{linear}_{m,2,D} $$
Here, $R^{linear}_{m,2,D}$ is the Rademacher complexity of $D$-dimensional linear predictors with unit $\ell_2$ norm with respect to a set of $m$ samples. Since $p=2$, by the same Corollary 2  \cite{pmlr-v40-Neyshabur15}, we have a bound for this term: $R^{linear}_{m,2,D} \le \sqrt{\frac{{\min{\{p^*,4log(2D)}\} \max_i \|x_i\|^2_{p^*}}}{m}}$.
Let $K = \frac{2^d}{d^{d/2}} \sqrt{\frac{{\min{\{p^*,4log(2D)}\} \max_i \|x_i\|^2_{p^*}}}{m}}$. This $K$ is a positive constant that depends only on the fixed architecture ($d$), input dimensionality ($D$), sample size ($m$), and the maximum $\ell_2$ norm of input data points (which is assumed finite).

So, the Rademacher complexity bound for our class becomes:
$$ R_m(N_{d,H,\sigma_{RELU}}^{\mu_{2,2} \le \mu}) \le K \cdot \mu^d $$
For network $M_{high}$, its function $f_{W_{high}}$ belongs to the class  $\mathcal{F}_{\mu_{M_{high}}}$, and its Rademacher complexity is bounded by:
$$ R_m(\mathcal{F}_{\mu_{M_{high}}}) \le K \cdot \mu_{M_{high}}^d $$
For network $M_{low}$, its function $f_{W_{low}}$ belongs to the class  $\mathcal{F}_{\mu_{M_{low}}}$, and its Rademacher complexity is bounded by:
$$ R_m(\mathcal{F}_{\mu_{M_{low}}}) \le K \cdot \mu_{M_{low}}^d $$
As explained in Appendix 1.2, since we are given $\mu_{M_{low}}>\mu_{M_{high}}$, and $\mu_M, \mu_{M'}$ are non-negative (being norms), it directly follows that $\mu^d_{M_{low}}>\mu^d_{M_{high}}$.
Therefore:
\begin{equation} \label{eq:bound}
R_m(\mathcal{F}_{\mu_{M_{high}}}) \le K \cdot \mu_{M_{high}}^d <  K \cdot \mu_{M_{low}}^d \end{equation}
This shows that the Rademacher complexity of the hypothesis class associated with $M_{high}$ is strictly smaller than that associated with $M_{low}$.

From the generalization bound established in Appendix~\ref{appendix:theoretical_found}:
For $M_{high}$, with probability at least $1-\delta$:
$$ R_\mathcal{D}(f_{W_{high}}) - R_S(f_{W_{high}}) \le 2 R_m(\mathcal{F}_{\mu_{M_{high}}}) + C \sqrt{\frac{\log(1/\delta)}{m}} $$
For network $M_low$, with probability at least $1-\delta$:
$$ R_\mathcal{D}(f_{W_{low}}) - R_S(f_{W_{low}}) \le 2 R_m(\mathcal{F}_{\mu_{M_{low}}}) + C \sqrt{\frac{\log(1/\delta)}{m}} $$
Since \eqref{eq:bound}, it implies that the upper bound on the generalization gap for $M_{high}$ is lower than that for $M_{low}$, that is, in the worst case, the generalization gap $M_{high}$ can have is strictly less than $M_{low}$. In other words, $M_{high}$ is generally better generalized. We also provide a more intuitive interpretation of DON and NOD metrics in Appendix~\ref{appendix:intuition}.

\subsection{Intuition Behind DON and NOD Metrics}\label{appendix:intuition}
\textbf{DON as a Proxy for Generalization:}
A negative DON indicates that fine-tuning on \( D_i \) increases the model’s weight magnitude. As shown above, we have proved that the model with high DON lead to better generalization. This is also supported by the study of \cite{10.5555/2621980}, its increment relates to the regularization principles, where smaller norms often correlate with lower generalization error (e.g., weight decay). Intuitively, samples that induce a significant increase in the Frobenius norm contribute to the complexity of the model, potentially damage its ability to generalize across domains, and high and positive DON indicate the simplification of the model and better generalization. Our experimental results support this intuition, showing that samples yielding high DON positive values improve cross-domain accuracy.

\textbf{NOD as an Indicator of Bad Sample:}
A high NOD value indicates that the data point $D_i$ has a significant influence on the model. In the context of ad-hoc SFT, training begins with a model that already possesses a certain level of generalization. Since the model is unlikely to encounter entirely new information or learn fundamentally new concepts after pretraining, data points with high NOD values are often indicative of low-quality samples, such as mislabeled or noisy data. Therefore, our method focuses on filtering out samples with high NOD values, thereby removing noisy or unlearnable data from the training process.
\subsection{Choice of Frobenius Norm}
The Frobenius norm is a matrix norm defined for a matrix \( W \in \mathbb{R}^{m \times n} \) as the square root of the sum of the squares of its elements, i.e., $\| W \|_F = \sqrt{\sum_{i=1}^m \sum_{j=1}^n |w_{i,j}|^2}$.
Compared to other norms, it offers specific advantages. For instance, the $\ell_1$ norm is given by $\| W \|_1 = {\sum_{i=1}^m \sum_{j=1}^n |w_{i,j}|}$, it treats the matrix as a flattened vector and measures the total absolute deviation. 
The $\ell_1$ norm’s robustness to outliers makes it suitable for measuring aggregate influence.
However, it lacks sensitivity to the fine-grained linear transformation differences represented by the weight matrix, which is critical for data pruning. 
Moreover, it provides a less effective measure of the magnitude difference for fine-grained data pruning. To capture the slightest difference between samples, the Frobenius norm shows better performance. 
In terms of the $\ell_2$ norm, for a matrix, it typically implies the Spectral norm and is given by $\| W \|_2 = \sigma_{max}(W)$,
where $\sigma_{max}(W)$ is the maximum singular value obtained from the singular value decomposition (SVD) of $W$. This makes the $\ell_2$ norm solely focus on the largest singular value, capturing the dominant direction of the matrix's transformation but ignoring the contribution of smaller singular values. This ignorance of finer structural changes in the weight matrix, making it less suitable for detecting sample-wise influences. Furthermore, the computation of SVD for a large matrix, which is a common situation for modern LLMs, can be a heavy workload, hindering the scalability of the method. For the same matrix  \( W \in \mathbb{R}^{m \times n} \), comparing with the expensive computation of $\ell_2$ spectral norm ($O(min(mn^2,m^2n))$), Frobenius norm is much more efficient and only requiring $O(mn)$.

\subsection{Integration with TOPSIS} \label{topsis}
\begin{enumerate}
\item Normalization:  
   DON and NOD are normalized to eliminate scale differences. Given a matrix \( W \in \mathbb{R}^{n \times 2} \), where \( n \) is the number of samples and columns represent DON and NOD, vector normalization is applied:  
\begin{equation}
    \tilde{w}_{i,j} = \frac{w_{i,j}}{\sqrt{\sum_{k=1}^n w_{k,j}^2}}.
\end{equation}

\item Ideal Solutions:  
   The hypothetical ideal solution \( Z^+ \) and negative-ideal solution \( Z^- \) are defined as:  
\begin{equation}
     Z^+ = [\max(\tilde{W}_{\text{DON}}), \min(\tilde{W}_{\text{NOD}})]
\end{equation}
\begin{equation}
   Z^- = [\min(\tilde{W}_{\text{DON}}), \max(\tilde{W}_{\text{NOD}})]
\end{equation}
  
These represent the hypothetical "best" and "worst" cases, where DON is maximized and NOD minimized (for \( Z^+ \)), and vice versa (for \( Z^- \)).  

\item Distance Calculation:  
   The Euclidean distance of each sample to \( Z^+ \) and \( Z^- \) is computed:  
   \begin{equation}
     D_i^+ = \sqrt{\sum_{j=1}^2 (\tilde{w}_{i,j} - Z_j^+)^2} \quad  
   \end{equation}
   \begin{equation}
   D_i^- = \sqrt{\sum_{j=1}^2 (\tilde{w}_{i,j} - Z_j^-)^2}
   \end{equation}

\item TOPSIS Score:  
   The final score for each sample is:  
   \begin{equation}
       C_i = \frac{D_i^-}{D_i^+ + D_i^-},
   \end{equation}
   
   where \( C_i \in [0, 1] \). Higher scores indicate closer alignment with \( Z^+ \) and greater divergence from \( Z^- \).  
\end{enumerate}

\subsection{Training Settings}
\label{app:training}
\begin{itemize}
    \item Global batch size: 16
    \item Micro-batch size: 1
    \item Learning rate: 2e-5
    \item Optimizer: AdamW \cite{loshchilov2019decoupledweightdecayregularization}
    \item Warmup-ratio 0.025
    \item Every model is trained on 8 NVIDIA A100 GPUs for less than 1 hour
    
\end{itemize}

\subsection{Bad Samples Gallery} \label{badsamples}
\begin{lstlisting}[breaklines=true]
How many edges does a cube have?
I need to identify how many edges a cube has. I remember that a cube is a three-dimensional shape with six square faces.Each square face has four edges, but when considering edges that are shared between two faces, I can't count all edges separately.A cube has 12 unique edges. They can be grouped into three sets of four edges that run parallel to each of the cube's three dimensions.Thus, after considering all faces and avoiding double counting, I confirm that the total number of edges in a cube is 12.\n Therefore the final answer is A cube has 12 edges.

Find the number of vertices of the rectangular prism.
I need to find the number of vertices of a rectangular prism.I know that a rectangular prism is a 3D shape with rectangular faces, length, width, and height.I can imagine that each corner of the prism is a vertex, and there are 2 corners at each edge.I can count 4 vertices at the top face, 4 vertices at the bottom face, and no additional vertices at the edges, since each edge has 2 vertices already counted.I conclude that there are 4 + 4 = 8 vertices in a rectangular prism.\n Therefore the final answer is A rectangular prism has 8 vertices.

If you select three of the four numbers 1, 2, 3, and 4, allowing duplicates, and arrange them in a row to form a three-digit natural number. How many multiples of 3 are there?\n 
Therefore the final answer is There are 22 multiples of 3 that can be formed by selecting three of the four numbers 1, 2, 3, and 4 (allowing duplicates) and arranging them in a row to form a three-digit number.

What is the smallest number among the three numbers 5, 8, and 4?
I am asked to find the smallest number among three given numbers: 5, 8, and 4.I will compare the three numbers by looking at their values.I can see that 4 is the smallest number among the three, as it has the lowest value.\n Therefore the final answer is The smallest number among 5, 8, and 4 is 4.
 
The function $f$ is defined as follows: \n$$f(x) = 4 + \frac{\\text{sqrt}}{x}.$$ \nWhat is the value of $f(144)$?\n\nA. 0 \nB. 16 \nC. 40 \nD. 76
We start by substituting 144 into the function f(x). The function is defined as: f(x) = 4 + sqrt(x). So we need to calculate f(144).Substituting 144 into the function, we have: f(144) = 4 + sqrt(144).Next, we calculate sqrt(144). The square root of 144 is 12.Now we can substitute this back into the function: f(144) = 4 + 12.Adding these values together gives: f(144) = 16.\n Therefore the final answer is 16

You subtracted 4.3 when you should have added 4.3 to a number, resulting in 8.8. Find the correct value.
\n Therefore the final answer is The correct value is 17.4.

Karen bakes 50 chocolate chip cookies. She keeps 10 for herself, and she gives 8 to her grandparents. Karen wants to give everyone in her class cookies. Each person in her class will receive 2 cookies. How many people are in Karen's class?
Therefore the final answer is Karen has 16 people in her class.

Ronnie is a train engineer. His train pulls 3 red boxcars, 4 blue boxcars, and 7 black boxcars. The black boxcars can each hold 4000 pounds of coal, and the red boxcars can hold 3 times as much as the blue boxcars. All of the train's boxcars combined can hold 132,000 pounds of coal. What is the ratio of the coal capacity of each black boxcar to each blue boxcar?
I understand that Ronnie's train has 3 red boxcars, 4 blue boxcars, and 7 black boxcars, and we need to find the ratio of the coal capacity of each black boxcar to each blue boxcar.I recognize that the black boxcars can hold 4000 pounds of coal, and the red boxcars can hold 3 times as much as the blue boxcars, and the total capacity of all boxcars is 132,000 pounds.I will let the coal capacity of each blue boxcar be B pounds, so each red boxcar can hold 3B pounds.I can write an equation based on the total capacity of all boxcars: 3(3B) + 4B + 7(4000) = 132,000.I simplify the equation by combining like terms: 9B + 4B + 28,000 = 132,000.I combine like terms to get: 13B + 28,000 = 132,000.I subtract 28,000 from both sides to get: 13B = 104,000, and then divide both sides by 13 to get: B = 8,000.I know that each blue boxcar can hold 8,000 pounds, and each black boxcar can hold 4,000 pounds, so the ratio of the coal capacity of each black boxcar to each blue boxcar is 4000/8000 = 1/2.
\n Therefore the final answer is The ratio is 1:2.
\end{lstlisting}



\subsection{Empirical Validations of Output Layer Focus}
\label{app:validation}
\begin{figure}[H]
\begin{center} 
\includegraphics[width=\linewidth]{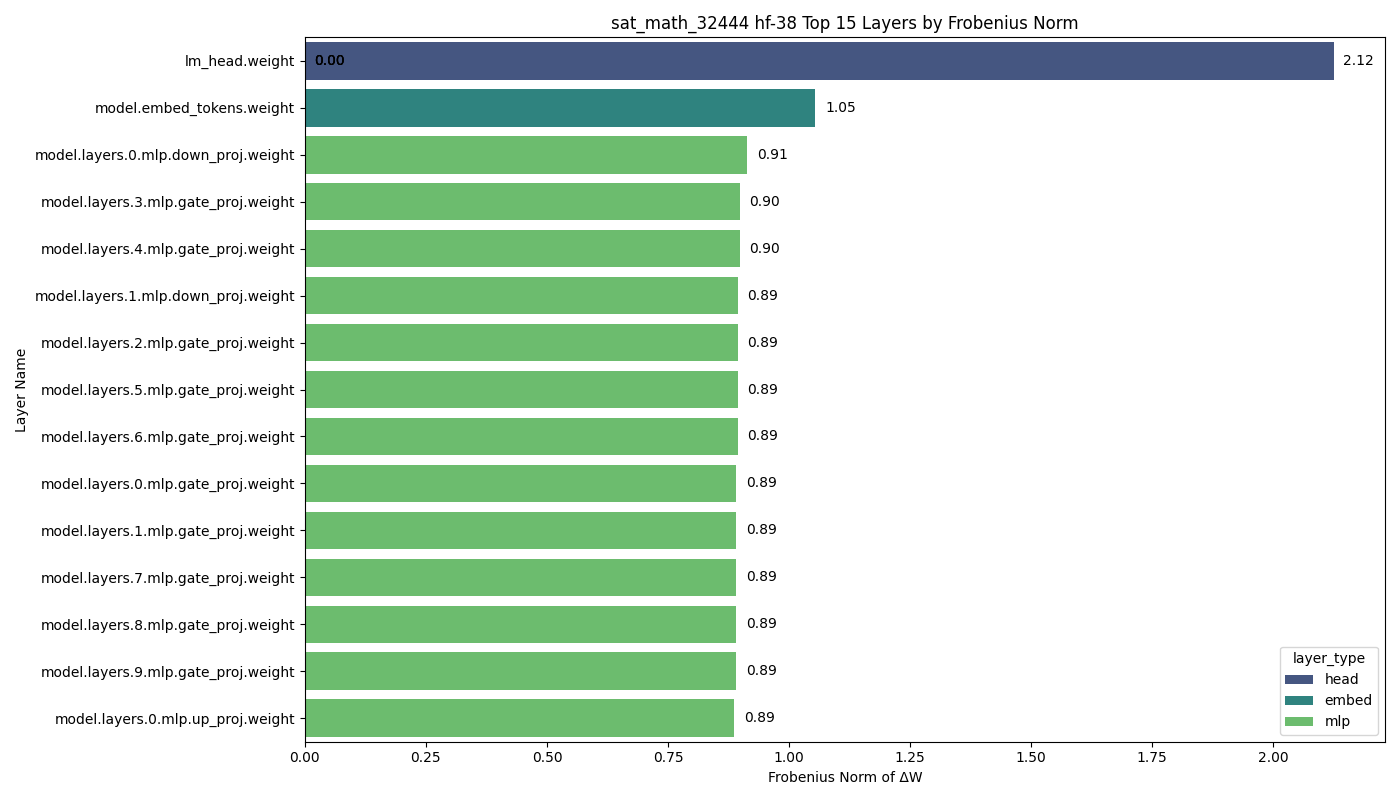}
    \caption{Ranking of Frobenius norm delta of layers of Llama-3.1-8B-Instruct after fine-tuning on SAT Math COT dataset, epoch 2} \label{fig:layer_rank}
\end{center}
\end{figure}

\begin{figure}[h!]
\begin{center} 
\includegraphics[width=\linewidth]{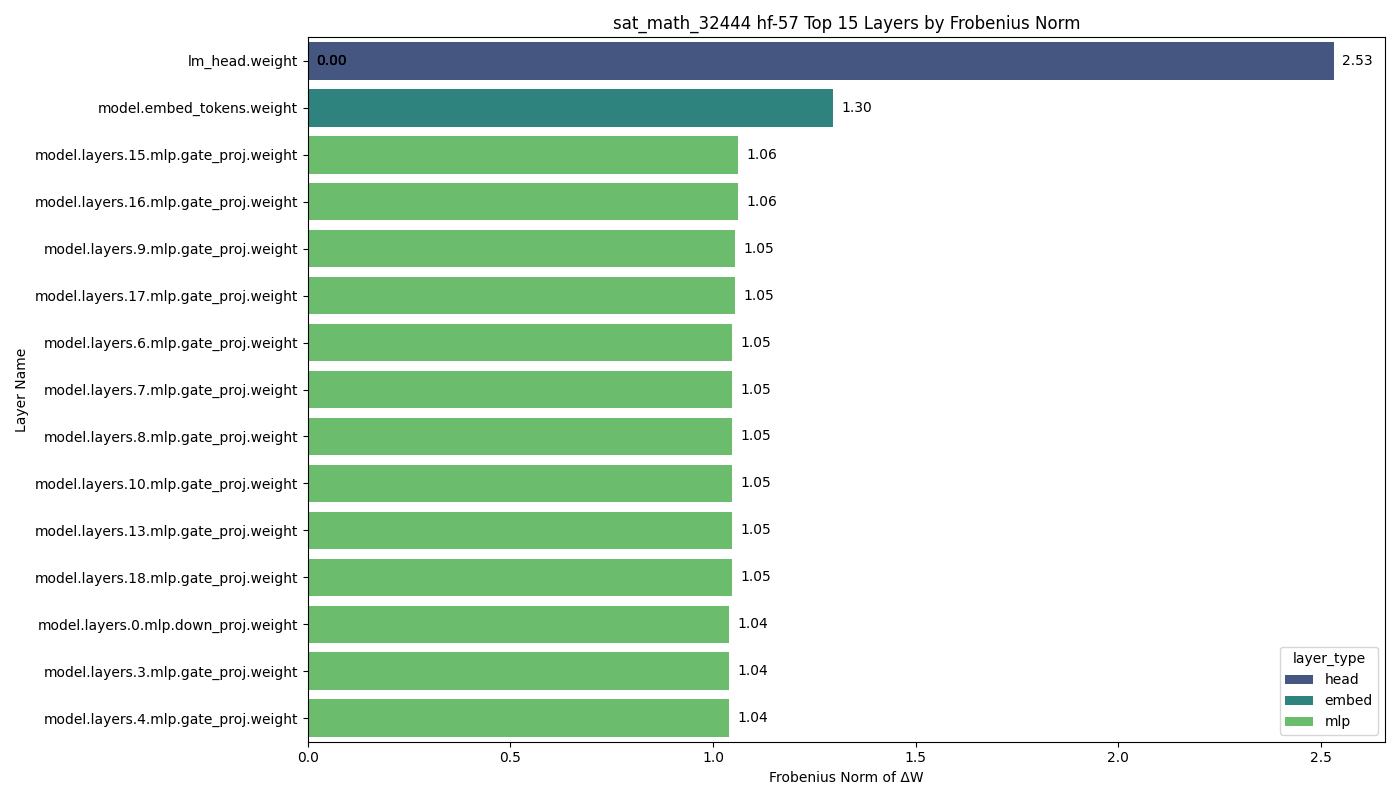}
    \caption{Ranking of Frobenius norm delta of layers of Llama-3.1-8B-Instruct after fine-tuning on SAT Math COT dataset, epoch 3} \label{fig:layer_rank_2}
\end{center}
\end{figure}

\begin{figure}[h!]
\begin{center} 
\includegraphics[width=\linewidth]{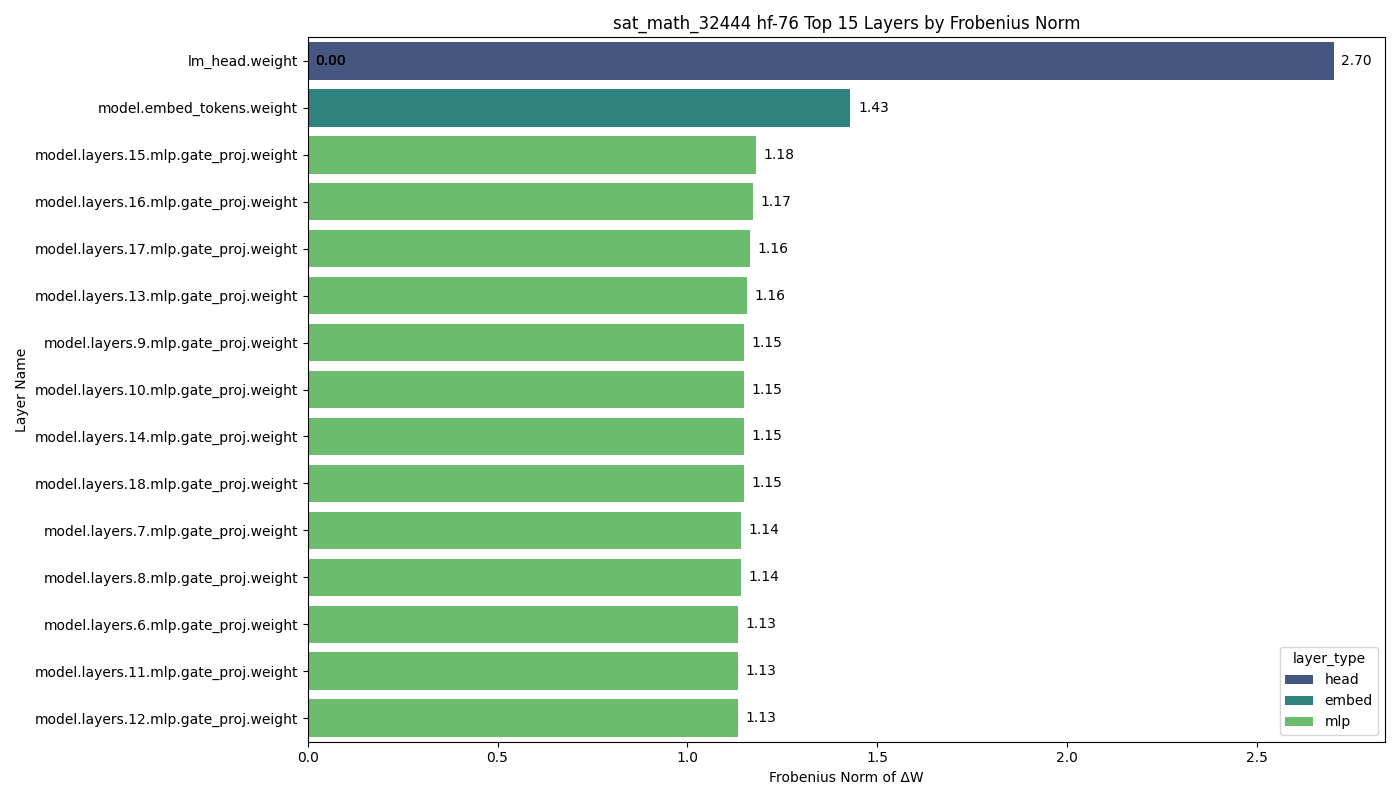}
    \caption{Ranking of Frobenius norm delta of layers of Llama-3.1-8B-Instruct after fine-tuning on SAT Math COT dataset, epoch 4} \label{fig:layer_rank_3}
\end{center}
\end{figure}

\begin{figure}[h!]
\begin{center} 
\includegraphics[width=\linewidth]{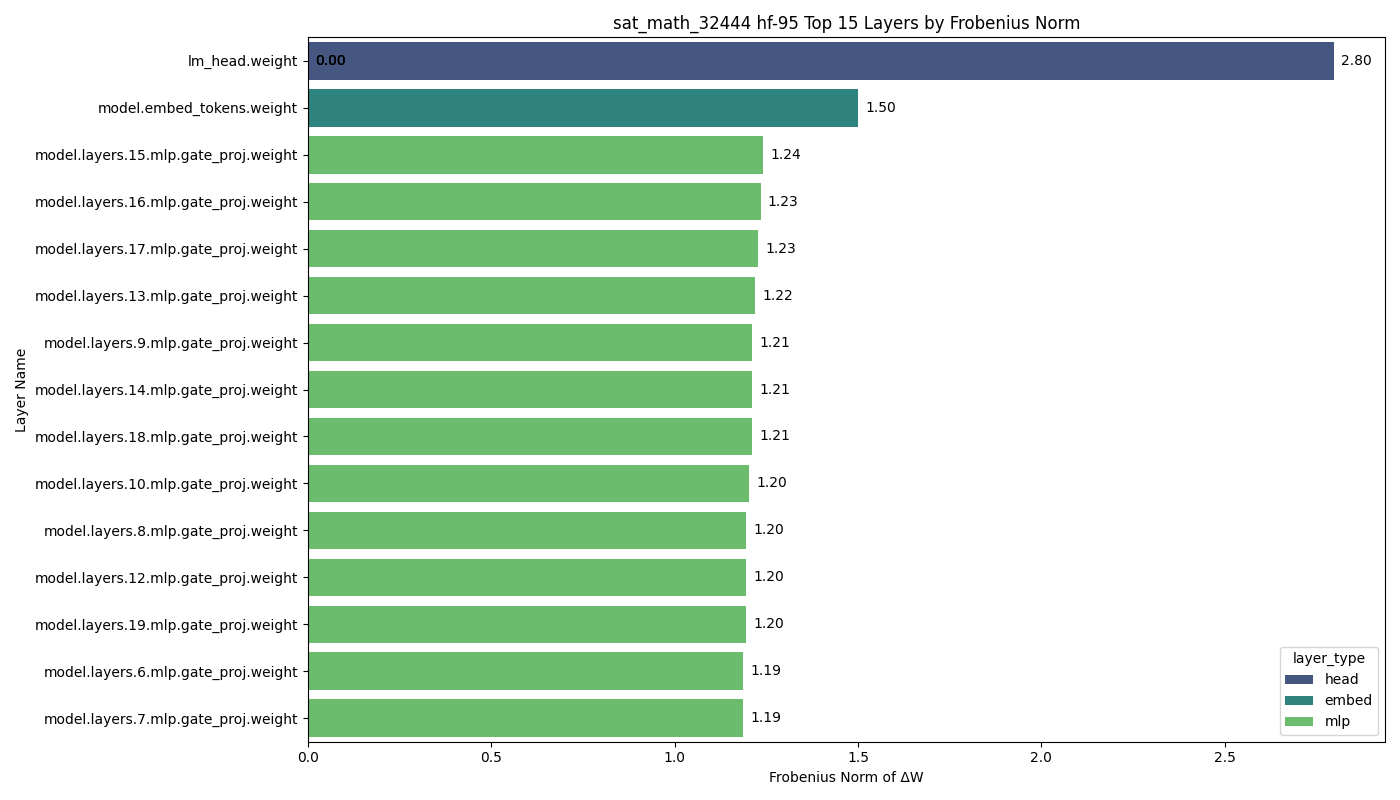}
    \caption{Ranking of Frobenius norm delta of layers of Llama-3.1-8B-Instruct after fine-tuning on SAT Math COT dataset, epoch 5} \label{fig:layer_rank_4}
\end{center}
\end{figure}

\begin{figure}[h!]
\begin{center} 
\includegraphics[width=\linewidth]{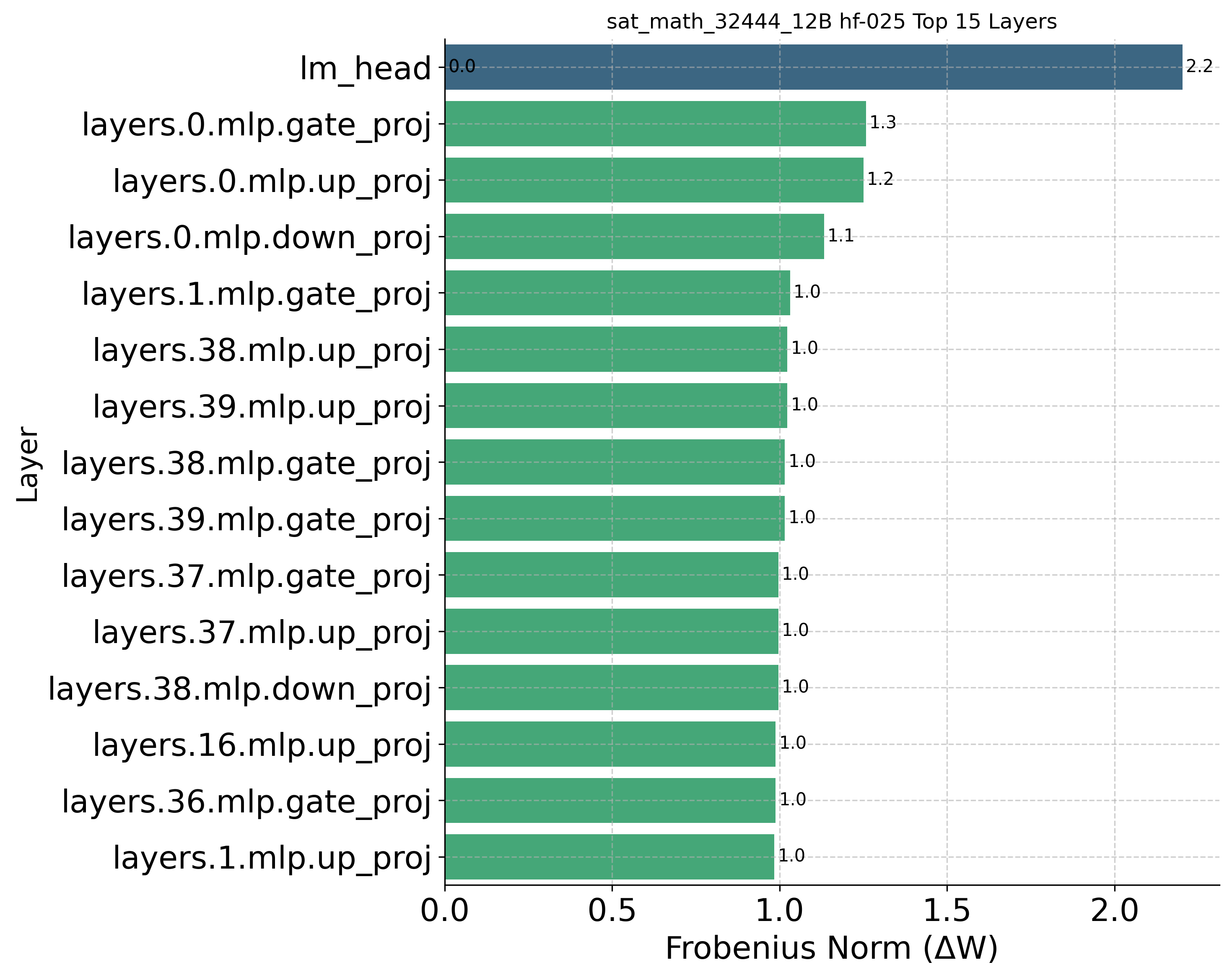}
    \caption{Ranking of Frobenius norm delta of layers of Llama-2-13b-chat after fine-tuning on SAT Math COT dataset, epoch 1} \label{fig:layer_rank_4}
\end{center}
\end{figure}

\begin{figure}[h!]
\begin{center} 
\includegraphics[width=\linewidth]{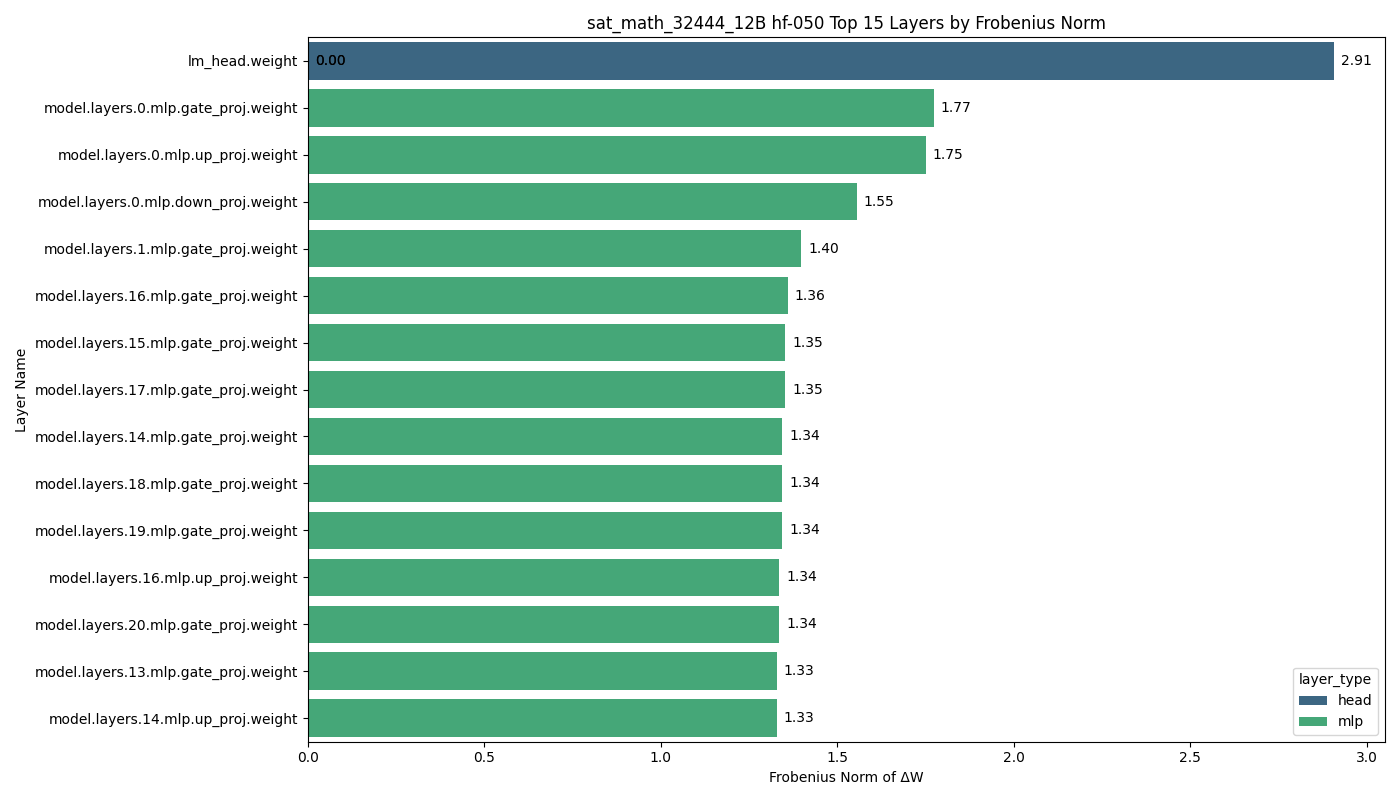}
    \caption{Ranking of Frobenius norm delta of layers of Llama-2-13b-chat after fine-tuning on SAT Math COT dataset, epoch 2} \label{fig:layer_rank_4}
\end{center}
\end{figure}

\onecolumn 
\subsection{Benchmark Details} \label{benchmark-table}
\begin{table*}[h!]
\begin{center}
\caption{Constructed Benchmark Details. To ensure the representativity of each measured ability, multiple datasets focusing on the same ability are selected, mitigating potential biases arising from the limited scope of individual exams. Additionally, the inclusion of the comprehensive and unrelated dataset IFEval-en further enhances the reliability and representative of the benchmark, aligning it more closely with real-world use cases.}
\begin{tabular}{llcc}
\toprule
\multicolumn{1}{c}{\bf Dataset}  & \multicolumn{1}{c}{\bf Ability} & \multicolumn{1}{c}{\bf Validation Set Available} & \multicolumn{1}{c}{\bf Test Set Size} \\
\midrule
LSAT-AR         & Logical Reasoning          & False & 230 \\
LSAT-LR         & Logical Reasoning          & False & 510 \\
LogiQA-en          & Logical Reasoning          & True  & 651 \\
LSAT-RC         & Reading Comprehension      & False & 269 \\
SAT-en          & Reading Comprehension      & False & 206 \\
AQUA-RAT        & Mathematical Problem-Solving & True  & 254 \\
SAT-math        & Mathematical Problem-Solving & False & 220 \\
IFEval-en       & Instruction Following      & False & 541 \\
\bottomrule
\end{tabular}
\end{center}
\end{table*}

\subsection{Datasets Details}
\begin{table*}[h!]
\begin{center}
\caption{Details of the datasets used in the study.}\label{datasets-table}
\begin{tabular}{llcc}
\toprule
\multicolumn{1}{c}{\bf Dataset}  & \multicolumn{1}{c}{\bf Ability} & \multicolumn{1}{c}{\bf Label Type} & \multicolumn{1}{c}{\bf Size} \\
\midrule
LogiQA Train           & Logic Reasoning            & Single letter only     & 7,851 \\
SAT Math COT & Mathematical Problem-Solving & COT & 32,444 \\
GSM8K & Mathematical Problem-Solving & COT & 8794 \\
Ifeval-Like Data & Instruction Following & General text & 56.3K \\
\bottomrule
\end{tabular}
\end{center}
\end{table*}

\subsection{Cross Architecture Results} \label{cross_results}


\begin{table*}[h!]
\centering
\small
\caption{Experimental results comparing different methods on Llama-3.2-3B-Instruct. The \textbf{Non-Target} column shows average performance excluding mathematical reasoning (target domain), revealing how methods generalize to other abilities. All values are percentages.}\label{tab:table1}
\begin{tabular}{lcccccc}
\toprule
\textbf{Method} & 
\multicolumn{1}{c}{\bf \makecell{Logic\\(↑)}} & 
\multicolumn{1}{c}{\bf \makecell{Reading\\(↑)}} & 
\multicolumn{1}{c}{\bf \makecell{Math\\(↑)}} & 
\multicolumn{1}{c}{\bf \makecell{IFEval\\(↑)}} & 
\multicolumn{1}{c}{\bf \makecell{Non-Target\\(↑)}} & 
\multicolumn{1}{c}{\bf \makecell{All\\Avg (↑)}} \\
\midrule
ALL (100\%)      & 9.23  & 34.67 & 10.60 & 63.96 & 35.95 & 23.10 \\
DONOD 20\%       & 27.65 & 50.46 & 34.78 & 65.80 & 47.97 & 38.71 \\
DONOD 30\%       & 29.46 & 56.75 & 33.92 & 65.06 & 50.42 & 41.03 \\
\bottomrule
\end{tabular}

\end{table*}


\subsection{Ablation Configurations}
\begin{table}[h]
\begin{center}
\caption{Ablation study configurations.}\label{ablation-table}
\begin{tabular}{lcccc}
\toprule
\multicolumn{1}{c}{\bf Configuration} & \multicolumn{1}{c}{\bf DON} & \multicolumn{1}{c}{\bf NOD} & \multicolumn{1}{c}{\bf TOPSIS} \\
\midrule
DON Only                    & $\checkmark$ & $\times$ & $\times$ \\
NOD Only                    & $\times$ & $\checkmark$ & $\times$ \\
DON + NOD (Weighted Sum)    & $\checkmark$ & $\checkmark$ & $\times$ \\
DON + NOD (Pareto Front)    & $\checkmark$ & $\checkmark$ & $\times$ \\
DON + NOD + TOPSIS (Full)   & $\checkmark$ & $\checkmark$ & $\checkmark$ \\
\bottomrule
\end{tabular}
\end{center}

\end{table}

\end{document}